\documentclass[letterpaper]{article} 
\usepackage{aaai2026}  
\usepackage{times}  
\usepackage{helvet}  
\usepackage{courier}  
\usepackage[hyphens]{url}  
\usepackage{graphicx} 
\usepackage{natbib}  
\usepackage{caption} 
\usepackage{algorithm}
\usepackage{algorithmic}

\usepackage{newfloat}
\usepackage{listings}
\DeclareCaptionStyle{ruled}{labelfont=normalfont,labelsep=colon,strut=off} 
\floatstyle{ruled}
\newfloat{listing}{tb}{lst}{}
\floatname{listing}{Listing}
\usepackage{amsmath}   
\usepackage{amssymb}   
\usepackage{amsfonts}
\usepackage{multirow}
\title{RadarMP: Motion Perception for 4D mmWave Radar in Autonomous Driving}

\author {
    Ruiqi Cheng\textsuperscript{\rm 1},
    Huijun Di\textsuperscript{\rm 1, \dag},
    Jian Li\textsuperscript{\rm 2, \rm 3, \dag},
    Feng Liu\textsuperscript{\rm 4},
    Wei Liang\textsuperscript{\rm 1}
}
\affiliations {
    
    \textsuperscript{\rm 1}School of Computer Science and Technology, Beijing Institute of Technology, Beijing, China\\
    \textsuperscript{\rm 2}Key Laboratory of Electronic and Information Technology in Satellite Navigation, Ministry of Education, Beijing, China\\
    \textsuperscript{\rm 3}Radar Technology Research Institute, School of Information and Electronic, Beijing Institute of Technology, Beijing, China\\
    \textsuperscript{\rm 4}Beijing Racobit Electronic Information Technology Co., Ltd., Beijing, China\\
    \textsuperscript{\dag}indicates corresponding authors \\
    chengrui7@bit.edu.cn, ~ajon@bit.edu.cn, ~lijian\_551@bit.edu.cn, ~liufeng@racobit.com, ~liangwei@bit.edu.cn
}

\begin{document}

\makeatletter
\def\copyright@on{F}
\makeatother

\maketitle

\begin{abstract}
Accurate 3D scene motion perception significantly enhances the safety and reliability of an autonomous driving system.
Benefiting from its all-weather operational capability and unique perceptual properties, 4D mmWave radar has emerged as an essential component in advanced autonomous driving.
However, sparse and noisy radar points often lead to imprecise motion perception, leaving autonomous vehicles with limited sensing capabilities when optical sensors degrade under adverse weather conditions.
In this paper, we propose RadarMP, a novel method for precise 3D scene motion perception using low-level radar echo signals from two consecutive frames. 
Unlike existing methods that separate radar target detection and motion estimation, RadarMP jointly models both tasks in a unified architecture, enabling consistent radar point cloud generation and pointwise 3D scene flow prediction.
Tailored to radar characteristics, we design specialized self-supervised loss functions guided by Doppler shifts and echo intensity, effectively supervising spatial and motion consistency without explicit annotations.
Extensive experiments on the public dataset demonstrate that RadarMP achieves reliable motion perception across diverse weather and illumination conditions, outperforming radar-based decoupled motion perception pipelines and enhancing perception capabilities for full-scenario autonomous driving systems.
\end{abstract}

 \begin{links}
     \link{Project:}{https://github.com/chengrui7/RadarMP}
 \end{links}

\section{Introduction}\label{intro}

Millimeter-wave (mmWave) radar plays a crucial role in autonomous driving perception and navigation \cite{hanqing_navi1, zan_navi2, jiaxin_navi3} systems due to its unique wavelength, which can penetrate weather obstacles. 
However, conventional CFAR-based radar detection methods \cite{gocfar,oscfar} rely on statistical assumptions and lack the capacity to model complex background clutter or dynamic scenes, resulting in degraded detection performance and producing sparse, noisy radar point clouds. 
Recent research has proposed deep learning methods \cite{rpdnet, mmemp, radelft} that utilize dense point clouds from LiDAR and camera to supervise radar target detection.  Due to their different electromagnetic characteristics, using optical sensors to train a radar point enhancement model forces radar to focus on some less prominent reflections in the heatmap and echo signals, thereby hindering the complementarity of multimodal sensing in autonomous systems \cite{mmradar_review}.

Precise 3D scene motion perception is essential for scene understanding in autonomous driving. Scene flow, which estimates the pointwise motion in the 3D world by leveraging two consecutive frames from cameras \cite{geonet, raftmsf} or LiDAR \cite{spflownet,msbrn,seflow} sensors, has been studied for years. 
With the emergence of 4D mmWave radar, the improved spatial resolution has made it feasible to perform the scene flow estimation task by radar data.
But radar point clouds produced by low signal-to-noise ratio (SNR) target detectors exhibit significant noise and temporal inconsistency, leading to relatively unreliable and inaccurate scene flow estimation.  
To the best of our knowledge, Ding et al. \cite{raflow, cmflow} employ self-supervision and cross-modal supervision to estimate scene flow between two frames of radar point clouds on the VoD dataset \cite{vod}, making them the only existing works in this context, and their performance remains substantially inferior to LiDAR-based methods.  

\begin{figure}
	\centering
	{\includegraphics[width=0.98\columnwidth, trim=0 0 0 0, clip, keepaspectratio]{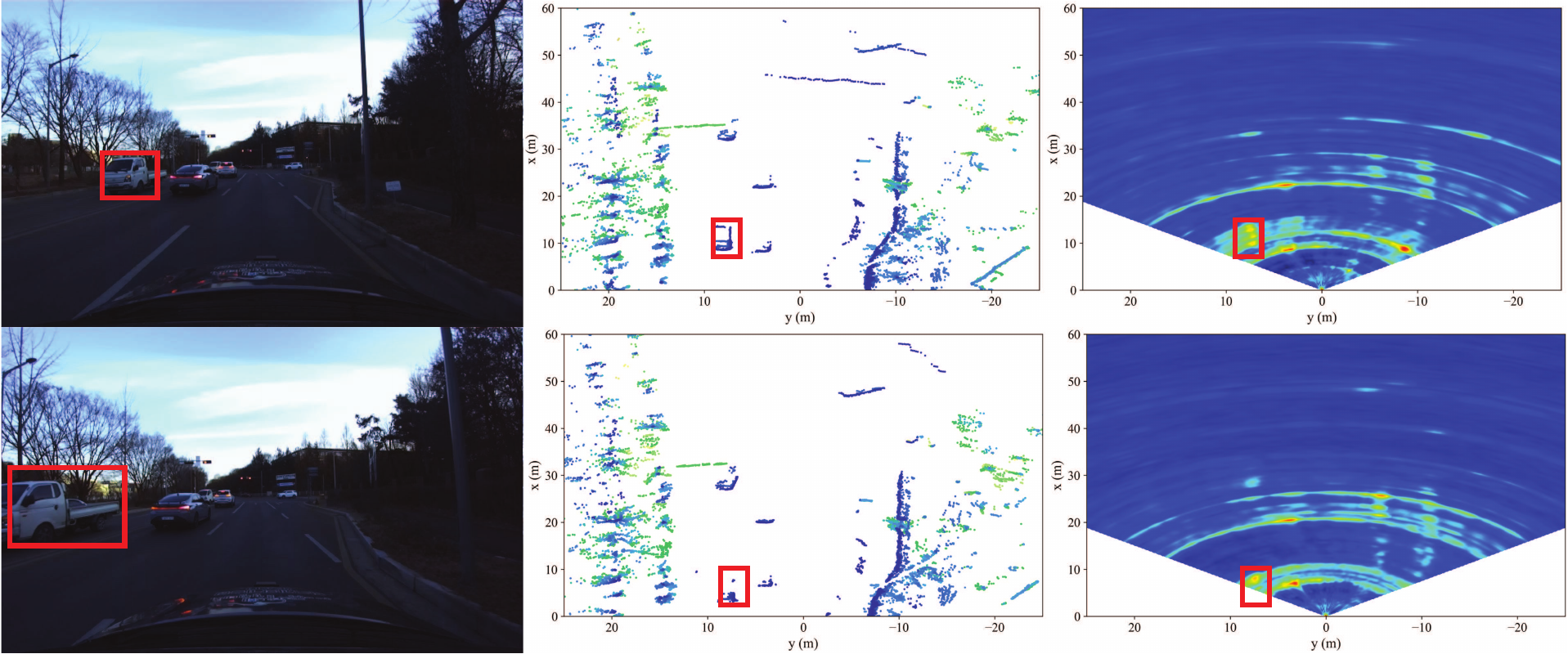}}
	\caption{\textbf{Motivation schematic.} The red boxes mark a target inter-frame motion in three modalities. The image and LiDAR are shown for visualization purposes only. In the radar heatmap, target motion aligns with the direction of energy propagation, revealing our key motivation.}
	\label{motivation}
\end{figure}
Our motivation stems from the assumption that the energy flow of target points across adjacent frames of radar echo signals should be consistent with the motion field, as shown in Figure~\ref{motivation}. In contrast, the energy flow of noise points tends to be disordered and irregular.
To enhance the scene motion perception capability of mmWave radar and remove false targets from radar echoes, we design a radar target detector that is consistent with the motion field of the adjacent radar frame, and simultaneously outputs scene flow estimation while detecting targets. 
Our method begins with the two successive 4D radar tensors, derived via multi-dimensional FFT, to perform consistent radar target detection and scene flow estimation. The radar tensor is a low-level representation of radar signals, storing echo intensity in a 4D cube (i.e., a tesseract).
Several works \cite{radarocc, rl3dod} have achieved high-accuracy object detection and occupancy prediction tasks using radar tesseract, which demonstrates the excellent performance of this signal format in mmWave radar applications.

Our contributions can be summarized as follows:
\begin{enumerate}
\item{We propose RadarMP, the first architecture that jointly addresses mmWave radar target detection and scene flow estimation tasks, using adjacent frame radar tesseract signal inputs to generate consistent radar point clouds and scene flow outputs.} 
\item{We introduce multiple specialized self-supervised loss functions based on the Doppler characteristics and echo intensity of radar signals to supervise both point cloud generation and scene flow estimation.}
\item{We conduct extensive experiments on the public dataset to validate the performance and effectiveness of the proposed method, which significantly enhances the motion perception capability of mmWave radar in full-scenario autonomous driving systems.}
\end{enumerate}

\section{Preliminary}
\subsection{Radar Tesseract Generation Workflow}
Radar transmits electromagnetic beams via its transmit (TX) antennas, which are reflected by targets and received by the receive (RX) antennas as echo signals. Most 4D mmWave radars use Frequency-Modulated Continuous Wave (FMCW) signals for transmission. A single frame of FMCW radar typically consists of multiple transmission cycles (i.e., chirp) where the signal frequency increases linearly over a short time within a chirp.
Each TX-RX antenna pair processes the echo signals in a radar frame through a mixer and an Analog-to-Digital converter (ADC), resulting in digital signals referred to as raw ADC data.
All raw ADC data are organized along the signal duration, chirp index, and antenna pair dimensions to form a 3D complex data cube, where the three axes correspond to fast time, slow time, and channel, respectively.

\begin{figure}
	\centering
	{\includegraphics[width=\columnwidth, trim=0 0 0 0, clip, keepaspectratio]{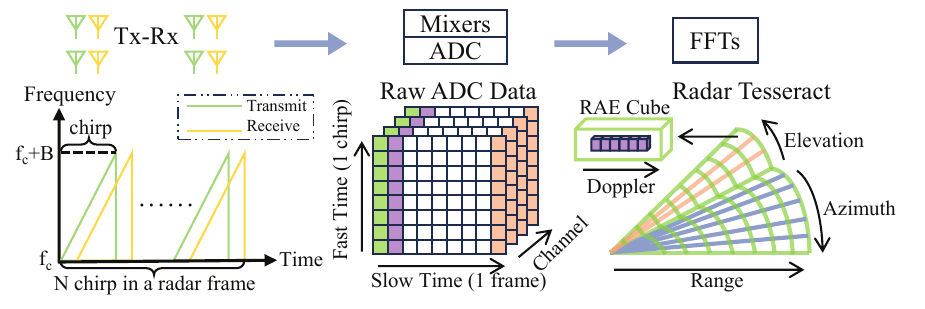}}
	\caption{\textbf{Tesseract generation pipeline}. Radar antenna array transmits multiple chirp signals per cycle. After receiving the echoes, the signals are mixed and sampled by ADCs to obtain the raw radar data, which is transformed into a radar tesseract via multi-dimensional FFT.}
	\label{tesseractworkflow}
\end{figure}

Fast Fourier Transforms (FFTs) are applied along the corresponding dimensions of the ADC data to extract detailed physical-domain information to construct a 4D radar tensor: range FFT recovers the propagation delay as range bins $r$, Doppler FFT estimates relative radial velocity $d$, and two spatial FFTs across the antenna array yield azimuth $a$ and elevation $e$ angles of arrival (AoA).
The detailed workflow is depicted in Figure~\ref{tesseractworkflow}.
In this paper, we refer to the 4D tensor as the tesseract, where each cell corresponds to the echo intensity at a location $(d, r, a, e)$ in the Doppler–range–azimuth–elevation space for a radar frame.

\subsection{Radar Tesseract for Motion Perception}
The radar tesseract exhibits a dense structure capable of capturing motion in complex 3D environments, combining the depth dimension of LiDAR with the dense coverage characteristic of camera images.
It preserves the comprehensive measurement of raw radar signals, avoiding the sparsity, noise, and clutter commonly introduced during traditional radar point cloud preprocessing.
For specific targets that are often overlooked during radar signal preprocessing but are critical in autonomous driving scenarios (such as pedestrians wearing low-reflectivity clothing, pets with fur, or asphalt road surfaces), the tesseract could significantly enhance motion perception and contribute to the safety and reliability of autonomous driving systems.

Despite its advantages, using the radar tesseract for motion perception remains a challenging task.
The dense structure of the radar tesseract results in substantial memory consumption (each frame in the Kradar dataset occupies nearly 300 MB), necessitating carefully designed models and processing strategies to mitigate GPU memory pressure and enhance computational efficiency.
Moreover, the noise within the tesseract is further amplified by the multipath effects inherent to mmWave signals, needing effective filtering mechanisms to suppress the adverse impact of such noise on motion perception performance.

\section{Methodology}
\subsection{Task Defination}
In this work, we address the motion perception task using two consecutive radar tesseracts output from a 4D mmWave radar. The input consists of a source frame $\mathbf{S} \in \mathbb{R}^{D \times R \times A \times E}$ and a target frame $\mathbf{T} \in \mathbb{R}^{D \times R \times A \times E}$, where $D$, $R$, $A$, and $E$ denote the Doppler, range, azimuth, and elevation dimensions respectively.

Our framework simultaneously solves two complementary subtasks:
\begin{enumerate}
\item{\textbf{Segmentation Prediction}: We generate a binary segmentation mask $\mathbf{M} \in \{0, 1\}^{R \times A \times E}$ for the source frame $\mathbf{S}$, where $\mathbf{M}(r,a,e) = 1$ identifies valid targets at spatial position $(r,a,e)$, while $\mathbf{M}(r,a,e) = 0$ denotes noise points.}
\item{\textbf{Scene Flow Estimation}: For each detected target point in the source frame $\mathbf{S}$ (where $\mathbf{M}(r,a,e) = 1$), we estimate a 3D scene flow field $\mathbf{F} = \{\mathbf{f}_i\}$. Each flow vector $\mathbf{f}_i = (\Delta r_i, \Delta a_i, \Delta e_i)$ represents the displacement of the target along the Range, Azimuth, and Elevation axes.}
\end{enumerate}

\subsection{Overview}
\begin{figure*}
	\centering
	{\includegraphics[width=\linewidth, trim=0 0 0 0, clip, keepaspectratio]{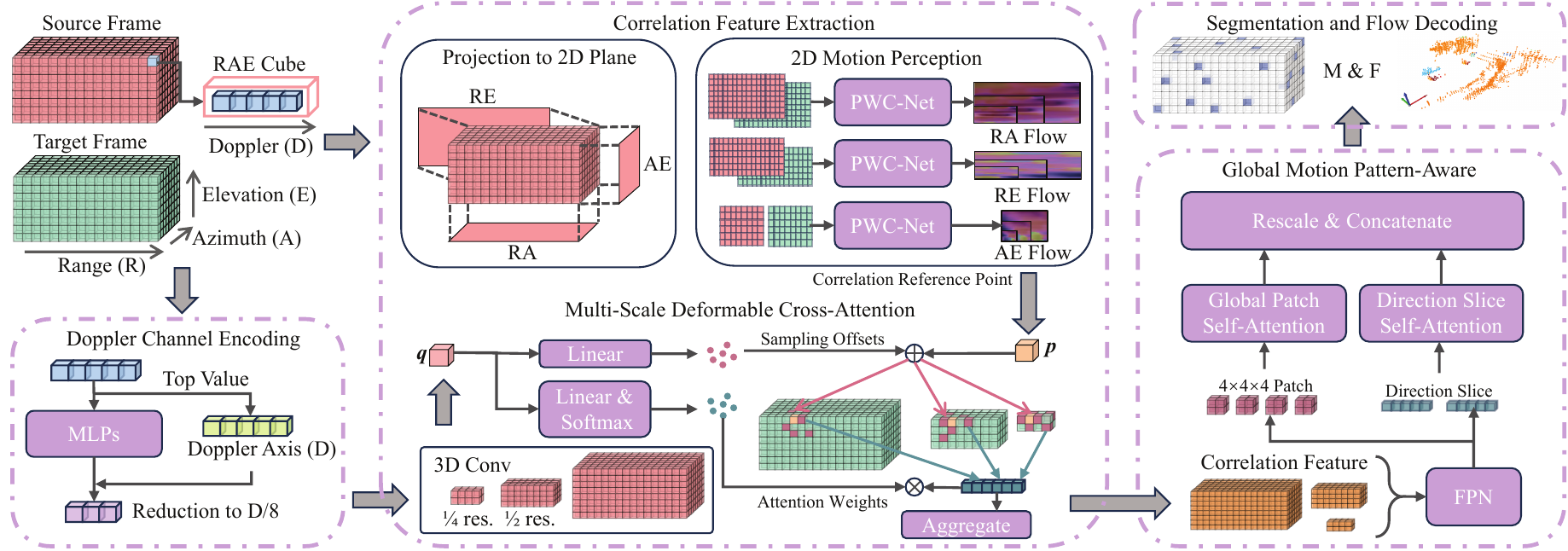}}
	\caption{\textbf{Pipeline Overview.} RadarMP processes two consecutive radar tesseracts through Doppler encoding and correlation feature extraction, followed by global motion pattern perception to derive motion cues, and finally decodes them into segmentation masks and flow predictions.}
	\label{pipeline}
\end{figure*}

The overall architecture is illustrated in Figure~\ref{pipeline}.
The two consecutive radar tesseracts are first unfolded along the three spatial planes, and a flow estimation network processes each plane to obtain coarse initial motion estimates and generate 3D reference points. After encoding along the Doppler channel dimension, the tesseracts are passed through dense 3D convolutions to extract multi-scale radar features. Combining the multi-scale features with the 3D reference points, RadarMP employs a multi-scale deformable cross-attention module to extract inter-frame correlation features. We enhance the correlation features with global context across different dimensions to distinguish motion cues, and finally decode to produce motion perception outputs trained in a self-supervised manner. The following section elaborates on the RadarMP framework and our tailored self-supervised loss functions.
\subsection{Doppler Channel Encoding}
Unlike the other three spatial dimensions, prior studies \cite{ekradar,rtnh,rl3dod} have considered the Doppler dimension $D$ at each $(r, a, e)$ coordinate to be redundant and often reduced to a minimal-dimensional representation by applying average or max pooling.
However, the Doppler axis at each spatial location encodes critical motion-related attributes, providing both semantic and physical cues for segmentation and scene flow estimation, relatively. 

To this end, it is necessary to encode the energy values along the Doppler dimension. We treat the Doppler axis in each voxel as feature channels and apply a multi-layer perceptron (MLP) to transform them into a compact representation. The encoded output is a Doppler-aware feature vector $\digamma_d \in \mathbb{R}^{C_d \times R \times A \times E}$.
In processing the Doppler axis, we not only consider the raw power, but also incorporate the corresponding Doppler velocity associated with each index.
The energy distribution along the Doppler axis reflects the confidence of each spatial location with respect to different Doppler velocities.
To capture this, we apply both Softmax and Gumbel-Softmax (for one-hot) \cite{gumbelsoftmax} functions to encode the Doppler velocity.
For each spatial location $(r,a,e)$, we denote the raw Doppler bins as $P_d\in\mathbb{R}^{D}$.
The Doppler velocity feature $\digamma_v$ is computed as follows:
\begin{equation}
\begin{aligned}
\digamma_{v1} &= \mathrm{sum}(\mathrm{matmul}(Ax_d, \mathrm{Softmax}(P_d)))\text{,}\\
\digamma_{v2} &= \mathrm{sum}(\mathrm{matmul}(Ax_d, \mathrm{GumbelSoftmax}(P_d)))\text{,}\\
\end{aligned}
\label{doppler}
\end{equation}
where $Ax_d\in\mathbb{R}^{D}$ denotes the Doppler axis.
Finally, we concatenate $\digamma_v$ with $\digamma_d$ forming the final Doppler-aware representation $\digamma_{dv} \in \mathbb{R}^{(C_d+2) \times R \times A \times E}$.

Note that the Doppler encoding process is applied identically to both the source frame $\mathbf{S}$ and target frame $\mathbf{T}$.
Consequently, this approach enables us to reduce the first dimension of tesseract to $D/8$ while retaining the essential characteristics of the Doppler bins.

\subsection{Correlation Feature Extraction}
Directly applying dense correlation over the entire 3D radar spherical space, as in image-based cost volumes, would lead to severe memory overhead.
Deformable attention \cite{deformdetr}  has demonstrated strong performance in 3D object detection and occupancy prediction tasks, while significantly reducing the computational complexity of attention modules.
Inspired by this, we treat the source frame $\mathbf{S}$ as the query and the target frame $\mathbf{T}$ as the value, and compute a correlation field using cross-deformable attention between the two radar tesseracts.

As a definition, given a query feature $\mathbf{q}$ and its corresponding reference point $\mathbf{p}$, deformable attention updates the query by aggregating features from the value feature $\mathbf{V}$ according to the following equation:
\begin{equation}
\begin{aligned}
&\mathrm{DeformAttn}(\mathbf{q}, \mathbf{p}, \mathbf{V}) = \\
&\sum_{m=1}^{M} W_m \left[ \sum_{k=1}^{K} A_{mk} \cdot W'_m\mathbf{V}(\mathbf{p} + \Delta \mathbf{p}_{mk}) \right]\text{,}
\end{aligned}
\label{deformattn}
\end{equation}
where  $\Delta \mathbf{p}_{mk}$ and $A_{mk}$ are learnable sampling offsets and learnable attention weight predicted from the query $\mathbf{q}$ for its $m_{th}$ head and $k_{th}$ sampling point, $\mathbf{V}(\mathbf{p} + \Delta \mathbf{p}_{mk})$ is the value features at the sample location $(\mathbf{p} + \Delta \mathbf{p}_{mk})$, and $W_m$ and $W'_m$ are the learnable transformation matrix for the $m_{th}$ attention head.

\subsubsection{Correlation Reference Point}
For optimal correlation feature capture through deformable attention, we should align the reference points as closely as possible to their actual warped locations in the target frame.
Inspired by the projection of 3D cubes onto 2D planes in DPFT \cite{dpft}, we project both $\mathbf{S}$ and $\mathbf{T}$ onto range-azimuth (RA), range-elevation (RE), and elevation-azimuth (AE) planes. Subsequently, we employ a pretrained PWC-Net \cite{pwcnet} architecture to predict the energy flow directions on these 2D projections. 
We slightly modify the original PWC-Net to regress 2D motion fields at three different scales (i.e., 1, 1/2, and 1/4) for providing multi-scale reference point.
The output motion fields consist of three 2D flow components: $\mathbf{F}_{ra}$, $\mathbf{F}_{re}$, and $\mathbf{F}_{ae}$.
Taking the RA-plane flow as an example, it is defined as $\mathbf{F}_{ra} = \{f^l_{ra}\}$, where each $f^l_{ra} \in \mathbb{R}^{2 \times \frac{R}{2^l} \times \frac{A}{2^l}}$, with $l = 0, 1, 2$.

The reference point coordinates $\mathbf{P}=\{\mathbf{p}_l\}$ at each feature level $l$ are computed by averaging the 2D flow predictions across three spatial planes (RA, RE, and AE). Specifically, the estimated motion in each plane is extended along the missing spatial dimension to construct full 3D flow volumes. The reference coordinates $\mathbf{p}_l\in \mathbb{R}^{3 \times \frac{R}{2^l} \times \frac{A}{2^l} \times \frac{E}{2^l}}$ are obtained by applying the averaged flow displacements to the original query grid positions along the range, azimuth, and elevation axes.

\subsubsection{Multi-Scale Deformable Cross Attention}
To capture motion information at multiple scales, we employ multi-scale deformable attention \cite{deformdetr, voxformer} to update the correlation features between $\mathbf{S}$ and $\mathbf{T}$.
First, we extract two three-level feature pyramids, $\digamma^{\mathbf{S}}_L$ and $\digamma^{\mathbf{T}}_L$, from the Doppler-encoded features $\digamma_{dv}^{\mathbf{S}}$ and $\digamma_{dv}^{\mathbf{T}}$ using a ResNet3D backbone \cite{resnet3d}. 
Taking the source frame as an example, the pyramid is defined as $\digamma^{\mathbf{S}}_L=\{\digamma_l^{\mathbf{S}}\}$, where each $\digamma_l^{\mathbf{S}} \in \mathbb{R}^{C_l \times \frac{R}{2^l} \times \frac{A}{2^l} \times \frac{E}{2^l}}$, with $l = 0, 1, 2$.
We apply the MLP to unify the channel dimensions across multiple scales and flatten them, producing the query $\mathbf{Q}_{\mathbf{S}}$ and the value $\mathbf{V}_{\mathbf{T}}$.
In our method, the correlation feature $\digamma_c$ between the two radar tesseracts is computed via the following multi-scale deformable attention equation:
\begin{equation}
\begin{aligned}
&\mathrm{MSDeformAttn}(\mathbf{q}, \mathbf{p}, \{\mathbf{v}_{\mathbf{T}}^{l}\}) = \\
&\sum_{m=1}^{M} W_m \left[ \sum_{l=1}^{L} \sum_{k=1}^{K} A_{mlk} \cdot W'_m\mathbf{v}_{\mathbf{T}}^{l}(\mathbf{p} + \Delta \mathbf{p}_{mlk}) \right] \text{,}
\end{aligned}
\label{msdeformattn}
\end{equation}
where $\mathbf{V}_{\mathbf{T}} = \{\mathbf{v}_{\mathbf{T}}^{l}\}$ is the multi-scale feature maps of the target frame, $\Delta \mathbf{p}_{mlk}$ and $A_{mlk}$ are the learnable sampling offsets and learnable attention weight predicted from the query $\mathbf{q}$ for its $k_{th}$ sampling point at the $l_{th}$ feature level and the $m_{th}$ head. 

Both the key and query elements are from the flatten multi-scale feature maps. The reference point $\mathbf{p} \in \mathbf{P}=\{p_l\}$ for each query feature $\mathbf{q} \in \mathbf{Q}_{\mathbf{S}}=\{\mathbf{q}_{\mathbf{S}}^{l}\}$ is computed at the corresponding scale via the pseudo warping operation defined in Eq.~\eqref{warprefpoint} (in Supp.).
By applying Eq.~\eqref{msdeformattn} to perform multi-scale feature interaction on $\digamma_{dv}^{\mathbf{S}}$ and $\digamma_{dv}^{\mathbf{T}}$, the resulting multi-scale correlation feature map $\digamma^{\mathbf{C}}_L=\{\digamma_l^{\mathbf{C}}\}$ is derived as follows:
\begin{equation}
\digamma^{\mathbf{C}}_l = \mathrm{MSDeformAttn}\bigl(\mathbf{q}, \mathbf{p}, \{\mathbf{v}_{\mathbf{T}}^{l}\}\bigr)\text{,} \quad \mathbf{q} \in \mathbf{q}_{\mathbf{S}}^{l}\text{.}
\end{equation}
By applying a Feature Pyramid Network (FPN) \cite{fpn} to aggregate the multi-scale correlation features, we obtain the correlation representation $\digamma_c \in \mathbb{R}^{C_c \times R \times A \times E}$, which captures the correspondence between the two radar frames across multiple resolutions.

\subsection{Global Motion Pattern-Aware Module}
Decoding target segmentation requires awareness of global motion patterns to distinguish the spatial distribution of tesseract motion. The motion characteristics of noise, static targets, and dynamic targets are disordered, globally correlated, and locally correlated, respectively. To provide sufficient cues for accurate segmentation, we introduce two self-attention modules to capture global context and enhance effective target detection.

\subsubsection{Global Patch Self-Attention}
$\digamma_c$ is derived from the summation of multi-scale correlation feature map $\digamma^{\mathbf{C}}_L$ and the query feature $\mathbf{Q}_{\mathbf{S}}$ during output.
Consequently, $\digamma_c$ contains both context feature and correlation feature.
To minimize memory consumption, we employ the token reduction strategy in conjunction with ViT \cite{vit}. Specifically, $\digamma_c$ is partitioned into $4\times4\times4$ patches, with each patch treated as one token in the Transformer encoder.
Since the pointwise motion vectors are correlated with the spatial coordinates of the corresponding targets, we similarly convert the polar coordinates of all spatial elements in the radar tesseract into $4\times4\times4$ patches as the positional encoding for the corresponding feature tokens.

\subsubsection{Direction Slice Self-Attention}
The final target flow field is also correlated with the directional vector of each point target, which remains constant across all range bins within the same $(a,e)$ slice. 
To preserve fine-grained segmentation features that may be degraded by volumetric patching, we propose a slicing strategy along the AE plane. 
We treat all range bins at the $(a,e)$ of $\digamma_c$ as a token to the Transformer encoder and input the directional vector as the positional encoding for the corresponding $(a,e)$ token, thereby enhancing the fine-grained representation of global motion pattern-aware.

\subsection{Segmentation and Flow Decoding}
The features with segmentation cues obtained from the global patch self-attention and direction slice self-attention do not match the spatial dimensions of the original tesseract, and thus need to be rescaled to the original resolution.
Obtained from global patch self-attention, $\digamma_p \in \mathbb{R}^{C_p \times \frac{R}{4} \times \frac{A}{4} \times \frac{E}{4}}$ is restored by rearranging the patches along the channel dimension, resulting in $\digamma'_p \in \mathbb{R}^{\frac{C_p}{64} \times R \times A \times E}$.
$\digamma_s \in \mathbb{R}^{R \times A \times E}$ is the output of the direction slice self-attention, whose spatial dimensions are compressed along the range axis, leaving only azimuth and elevation. We set its channel dimension equal to the number of range bins, so that each channel corresponds to the feature of one range bin.
We expand the channel dimension of $\digamma_s$ to 1 and concatenate it with $\digamma'_p$ to obtain the complete global motion pattern feature $\digamma_g \in \mathbb{R}^{C_g \times R \times A \times E}$.

The global motion pattern feature $\digamma_g$ and the correlation feature $\digamma_c$, which contains the source frame's contextual information, are respectively passed through MLPs and then concatenated to form a unified feature $\digamma_u$. This unified feature is then fed into the segmentation head and the flow head to produce the final outputs: the binary segmentation confidence $\mathbf{M}_s \in \mathbb{R}^{1 \times R \times A \times E}$ and the scene flow prediction $\mathbf{F}_s \in \mathbb{R}^{3 \times R \times A \times E}$, where each value in $\mathbf{M}_s \in [0,1]$.

\subsection{Loss Function}
To supervise motion perception, we introduce three self-supervised loss terms: segmentation energy loss $\mathcal{L}_{se}$, energy flow loss $\mathcal{L}_{ef}$, and radial flow segmentation loss $\mathcal{L}_{rfs}$.
The overall loss function is formulated as:
\begin{equation}
\mathcal{L} = \mathcal{L}_{se} +\mathcal{L}_{ef} + \mathcal{L}_{rfs}\text{.}
\end{equation}
This loss function jointly optimizes the network from three aspects: energy distribution, energy flow direction, and the interaction between energy flow and energy distribution on the Doppler channel, respectively.
\subsubsection{Segmentation Energy Loss}
We apply the Gumbel-Softmax and mean operations across the Doppler dimension of the tesseract to obtain the maximum and summation energy features, respectively. These are concatenated to form $E_f$, which represents the maximum energy characteristics at each $(r,a,e)$ coordinate. Then, the noise energy level $\tau_f$ is estimated from the local and channel energy statistics. The greater the difference between $E_f$ and $\tau_f$, the higher the relative energy of the point, indicating a higher likelihood of being a target candidate. Moreover, segmentation masks should exhibit consistency between the source and target frames. The segmentation energy loss is as follows:
\begin{equation}
\begin{aligned}
&\mathcal{L}_{se} = \mathbf{M}_s-\mathrm{sigmoid}(E_f^\mathbf{S}-\tau_f^\mathbf{S}) \\
&+\mathbf{M}_s \times ( \mathrm{warp} ( \mathbf{M}_s,\mathbf{F}_s)-\mathrm{sigmoid}(E_f^\mathbf{T}-\tau_f^\mathbf{T}))\text{.}
\end{aligned}
\end{equation}
\subsubsection{Energy Flow loss}
The flow field of a target point must align with its energy flow direction. Based on this assumption, we introduce the Energy Flow Loss. Unlike optical flow in images, the energy flow field also includes disordered noise flows. To mitigate the impact of such noise on the loss function, we use energy intensity as the weighting factor, encouraging the model to focus more on the flow directions of target points. 
The energy flow loss is as follows:
\begin{equation}
\mathcal{L}_{ef} = E_f^\mathbf{S} \times (E_f^\mathbf{S}-\mathrm{warp}(E_f^\mathbf{T},~\mathbf{F}_s))\text{.}
\label{efl}
\end{equation}
\subsubsection{Radial Flow Segmentation Loss}
The Doppler value (i.e., radial relative velocity) multiplied by the inter-frame time $\Delta t$ should approximate the radial projection of the truth flow at the target point, which is the core insight behind the self-supervision of the radial flow segmentation loss. 
We define the Doppler candidate values $\digamma_v$ (obtained from Doppler channel encoding, as shown in Eq.~\eqref{doppler}). For convenience, we convert the polar coordinates volume (obtained from the global patch self-attention) into Cartesian coordinates $C$. The directional vector volume $O$ (obtained from the direction slice self-attention) represents the direction vectors of grid centers relative to the radar.
The radial flow segmentation loss is as follows:
\begin{equation}
\begin{aligned}
&\delta_v=\digamma_v - \frac{\mathrm{warp}(C ,~\mathbf{F}_s) - C}{\Delta t}\odot O\text{,}\\
&\mathcal{L}_{rfs}= \mathbf{M}_s - \mathrm{sigmoid}(\alpha(\beta-\delta_v^2))\text{,}
\end{aligned}
\end{equation}
where $\alpha$ and $\beta$ represent the tolerance for $\delta_v$.

\section{Experiment}
\subsection{Experimental Setup}
\subsubsection{Dataset}
We conduct experiments on the K-Radar dataset \cite{kradar}, which is currently the only autonomous driving dataset that provides radar signals in the tesseract format. Moreover, the K-Radar dataset also includes time-synchronized multi-view images, LiDAR point clouds, odometry information, and annotations for 3D object detection and tracking. The front-view images and LiDAR point clouds enable comparison of the performance of different modalities in motion perception. We utilize the odometry data, along with the 3D detection and tracking annotations, to generate scene flow labels. 
\subsubsection{Implementation}
In our experiments, we train the model using the Adam optimizer \cite{adam}. The learning rate is initially set to 0.001 and decays exponentially by a factor of 0.9 every 2 epochs.
The multi-scale deformable cross-attention module in our method is configured with three attention heads and 50 sampling points. For the self-attention modules, we adopt the native PyTorch implementation with two heads and two layers.
We train RadarMP using three NVIDIA RTX 3090 GPUs. And RadarMP achieves 7.6 fps inference on a 3090 using
7.5 GB GPU memory. More details of RadarMP can be found in the appendix.

\begin{figure*}[ht]
	\centering
	{\includegraphics[width=\linewidth, trim=0 0 0 0, clip, keepaspectratio]{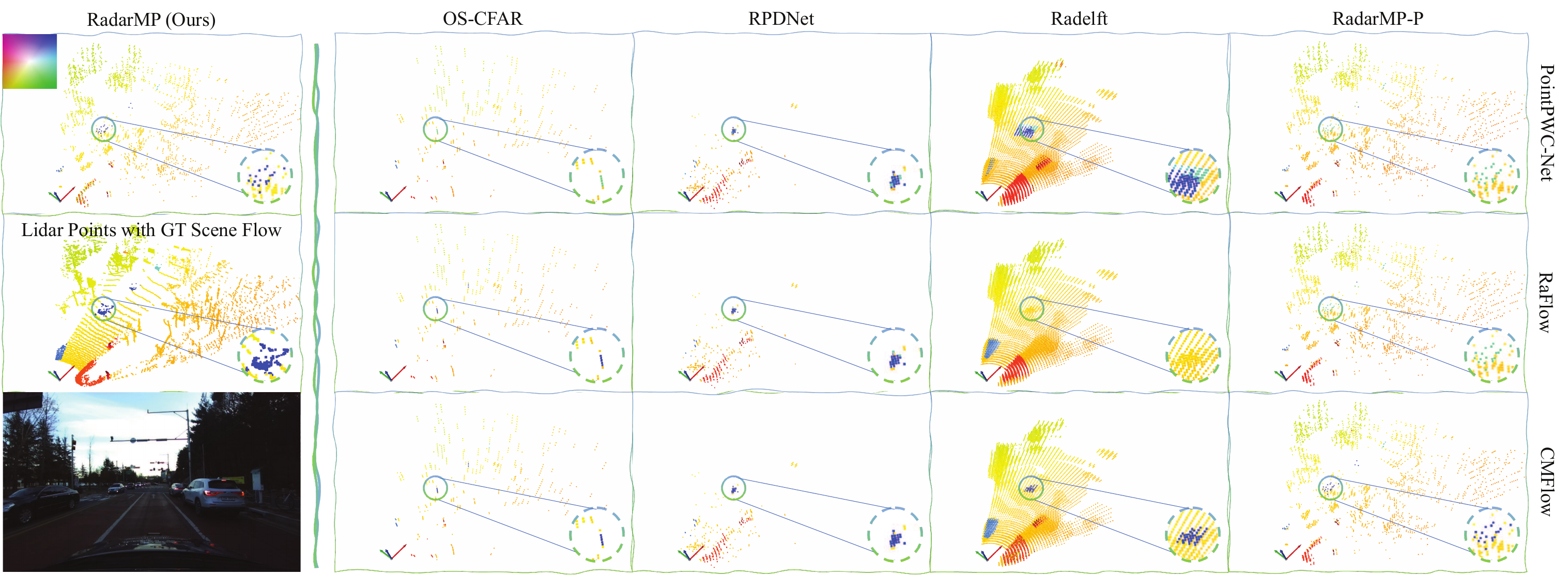}}
	\\[0.5em]  
	{\includegraphics[width=\linewidth, trim=0 0 0 0, clip, keepaspectratio]{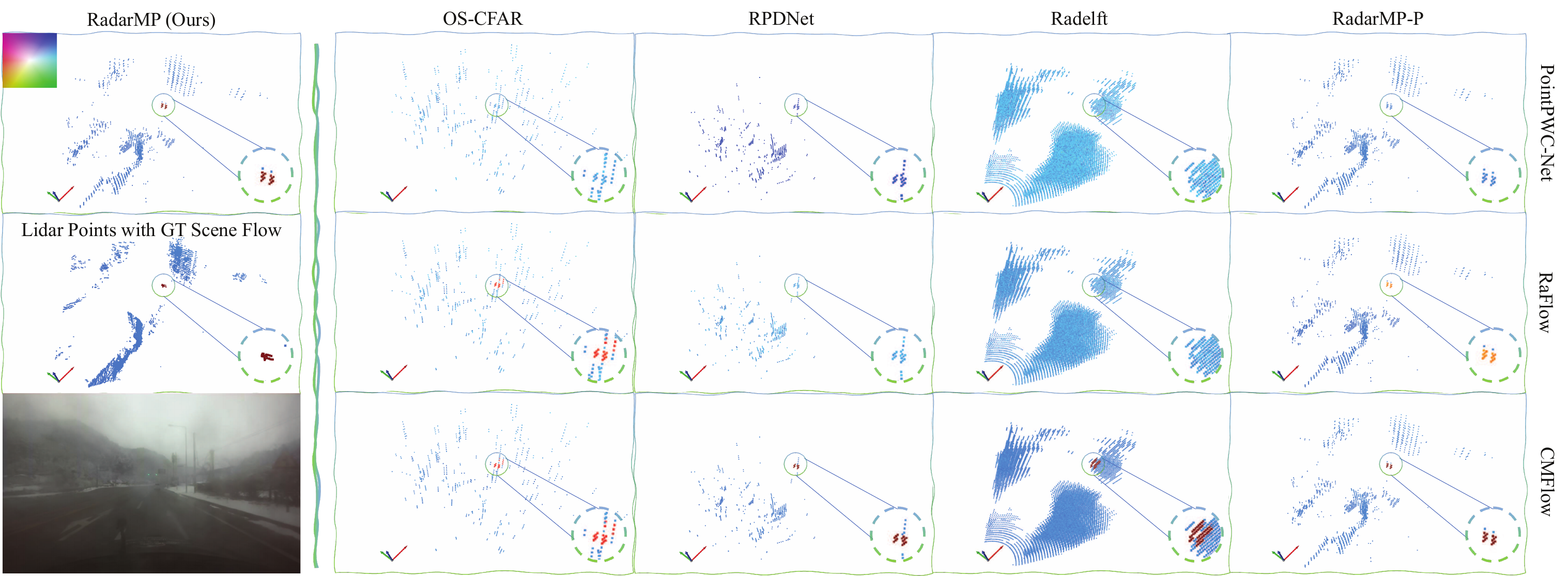}}
	\caption{\textbf{Qualitative results.}  The \textbf{left} side shows the motion perception output of RadarMP alongside LiDAR point clouds filtered by RoI with ground-truth scene flow. Columns 1–4 on the \textbf{right} side display radar target detection results from three segmentation baseline methods and our RadarMP (RadarMP-P), while rows 1–3 correspond to flow prediction results from different scene flow baselines. A dynamic object in the scene is zoomed in at the bottom right to highlight the accuracy of non-rigid motion estimation. Colors indicate motion vectors in the XY plane only and RGB image is used for visualization only.}
	\label{result}
\end{figure*}

\subsection{Segmentation Evaluation}
\subsubsection{Metrics}
To evaluate the segmentation performance between target and noise in the radar tesseract, we need to compare the occupancy consistency of the output against the LiDAR point cloud. 
When the output target point has at least 3 LiDAR points within its local neighborhood, we consider it a true positive. Based on this criterion, we compute probability of detection ($P_d$) and probability of False Alarm ($P_{fa}$) as segmentation metrics. We also calculate the Chamfer Distance (CD) between the predicted points and the voxel-filtered LiDAR point set to evaluate the spatial distribution of the target points.
Moreover, we calculate the average target energy and average noise energy based on the segmentation results to compute the signal-to-noise ratio (SNR) metric.
\subsubsection{Baselines}
For a comprehensive comparison, we select the traditional OS-CFAR method and two learning-based methods \cite{rpdnet, radelft} that enhance 3D radar point clouds using LiDAR supervision as baselines. In the OS-CFAR method, the number of background cells is set to 4, the guard cells to 1, and the false alarm rate to 1e-6. Moreover, we adopt the default hyperparameters for the learning-based methods to ensure fairness.
\subsubsection{Results}
We quantitatively compare with baseline methods on the test sets, as presented in Table~\ref{tableseg}. Compared to relying on energy thresholds or LiDAR supervision for segmentation, our method integrates energy constraints and scene motion consistency, resulting in a significantly mean higher detection probability of target points. Benefiting from the combination of multiple loss functions, RadarMP avoids introducing excessive pseudo targets, maintaining a low false alarm rate while achieving a clear perception of the surrounding scene, as illustrated in Figure~\ref{result}.
\begin{table}[!htbp]
	\caption{\textbf{Radar target detection results.} The \textbf{bold} number indicates the best result, and the \underline{underlined} number represents the second-best result. \textuparrow means bigger values are better, and vice versa.}
	\begin{center}
		\setlength{\tabcolsep}{6pt}
		\fontsize{9pt}{16pt}\selectfont
		\begin{tabular}{c|cccc}
			\hline
			\textbf{Method}&\textbf{$P_d$}(\%)\textuparrow&\textbf{$P_{fa}$}(\%)\textdownarrow&\textbf{CD(m)\textdownarrow}&\textbf{SNR(dB)\textuparrow}\\
			\hline
			\textbf{OS-CFAR}&1.643&\textbf{0.311}&10.030&\textbf{5.477}\\
			\textbf{RPDNet}&9.311&1.821&7.590&5.175\\
			\textbf{Radelft}&\underline{44.121}&6.200&\underline{6.553}&4.329\\
			\hline
			\textbf{RadarMP}&\textbf{69.458}&\underline{1.335}&\textbf{3.378}&\underline{5.232}\\
			\hline
		\end{tabular}          
		\label{tableseg}
	\end{center}
\end{table}
\subsection{Flow Evaluation}
\subsubsection{Metrics}
We adopt four commonly used scene flow metrics \cite{hpflownet} to evaluate the performance of flow field estimation: 1) EPE3D (m): the average end-point-error between the predicted and ground-truth scene flow of target points, 2) AccS3D (\%): the percentage of target points with endpoint error satisfying the strict condition (EPE3D $<$ 0.05 m or relative error $<$ 5\%), 3) AccR3D (\%): the percentage of target points meeting the relaxed condition (EPE3D $<$ 0.1 m or relative error $<$ 10\%), 4) Outlier3D (\%): the percentage of target points whose endpoint error exceeds the threshold (EPE3D $>$ 0.3 m or relative error $>$ 10\%).
\subsubsection{Baselines}
We selected two self-supervised scene flow estimation methods as flow evaluation baselines: PointPWC-Net (PPN), the first self-supervised method for point cloud scene flow estimation, and RaFlow\cite{raflow} and CMFlow\cite{cmflow}, the only two scene flow prediction models designed for mmWave radar point clouds supervised by self and cross-modal, respectively.
For a fair comparison, we apply these three methods to the radar points generated from the segmentation baseline, as well as the target point obtained from RadarMP (RadarMP-P), to predict scene flow. The scene flow estimated above is subsequently evaluated against the flow predictions generated by RadarMP.
\begin{table}[!htbp]
	\caption{\textbf{Scene flow evaluation results}. Baselines include combinations of a flow prediction model with different segmentation methods.}
	\begin{center}
		\setlength{\tabcolsep}{2.5pt}
		\fontsize{8pt}{12pt}\selectfont
		\begin{tabular}{c|c|cccc}
			\hline
			\multicolumn{2}{c}{\textbf{Method/Metric}}&EPE3D\textdownarrow&AccS3D\textuparrow&AccR3D\textuparrow&Outlier3D\textdownarrow\\
			\hline
			\multirow{4}{*}{\textbf{PPN}}&\textbf{OS-CFAR}&0.489&5.547&9.385&93.340\\
			&\textbf{RPDNet}&0.351&11.384&16.730&79.717\\
			&\textbf{Radelft}&0.496&6.337&18.669&89.060\\
			&\textbf{RadarMP-P}&0.223&20.005&39.741&59.624\\
			\hline
			\multirow{4}{*}{\textbf{RaFlow}}&\textbf{OS-CFAR}&0.329&11.635&20.887&82.399\\
			&\textbf{RPDNet}&0.311&12.900&25.650&78.122\\
			&\textbf{Radelft}&0.460&10.275&25.333&81.973\\
			&\textbf{RadarMP-P}&0.175&18.880&40.635&53.400\\
			\hline
			\multirow{4}{*}{\textbf{CMFlow}}&\textbf{OS-CFAR}&0.283&15.730&32.811&72.963\\
			&\textbf{RPDNet}&0.251&17.150&36.155&68.990\\
			&\textbf{Radelft}&0.190&20.151&46.584&65.263\\
			&\textbf{RadarMP-P}&\underline{0.168}&\underline{20.396}&\textbf{47.985}&\underline{50.841}\\
			\hline
			\multicolumn{2}{c|}{\textbf{RadarMP (Ours)}}&\textbf{0.157}&\textbf{21.365}&\underline{46.872}&\textbf{44.734}\\
			\hline
		\end{tabular}          
		\label{tableflow}
	\end{center}
\end{table}
\subsubsection{Results}
We show the evaluation metric results for flow prediction in Table~\ref{tableflow}. Taking the complementary self-supervised losses from segmentation and flow estimation, RadarMP predicts scene motion more accurately than baseline methods using only two consecutive radar tensors. Figure~\ref{result} presents qualitative comparisons with all baselines across two sample sequences. Compared to low-resolution and noisy point clouds, the motion of strong reflectors, such as vehicles, is more precisely captured within the tesseract. At the same time, global energy flow cues could infer rigid body motion. It can be observed that our predicted flow fields closely match the ground-truth motion.

\subsection{Ablation Study}
To validate the effectiveness of the three self-supervised loss functions we designed, we conduct ablation studies on different combinations of these losses. The results are shown in Table~\ref{tableablation}, where the bottom row reports the performance of RadarMP with the complete loss configuration. Each loss term contributes to improving the overall performance in both object detection and motion perception.
\begin{table}[!htbp]
\caption{\textbf{Ablation experiments on loss terms.} The $\surd$ indicates that the loss function is enabled in the model.}
\begin{center}
\setlength{\tabcolsep}{6pt}
\fontsize{9pt}{16pt}\selectfont
\begin{tabular}{ccc|ccc}
\hline
\textbf{$\mathcal{L}_{se}$}&\textbf{$\mathcal{L}_{ef}$}&\textbf{$\mathcal{L}_{rfs}$}&\textbf{$P_d$} (\%)\textuparrow&\textbf{$P_{fa}$}(\%)\textdownarrow&\textbf{EPE3D (m)\textdownarrow}\\
\hline
$\surd$&$\surd$&&62.033&2.258&0.209
\\
$\surd$&&$\surd$&56.224&3.847&0.788
\\
&$\surd$&$\surd$&19.846&17.136&0.621
\\
$\surd$&$\surd$&$\surd$&\textbf{69.458}&\textbf{1.335}&\textbf{0.157}
\\
\hline
\end{tabular}          
\label{tableablation}
\end{center}
\end{table}
To examine the dependence on PWCNet, which provides reference points for 2D flow predictions across three planes, the ablation study shows that turning off this module and using each cube’s own location as its reference point worsens the EPE3D metric by 0.31. When replacing PWCNet with a weak version, the EPE3D metric worsens by 0.07.

\section{Conclusion}
In summary, we propose RadarMP, a novel framework that leverages inter-frame energy propagation to enable motion perception for millimeter-wave radar in autonomous driving systems. RadarMP addresses two key tasks in an end-to-end manner: 1) target detection from low-level radar tesseract signals by performing motion-consistent segmentation of targets and noise in dense radar tensors, and 2) scene flow prediction for each segmented target point by extracting consecutive frame correlation features.
To retain radar sensing independence and complementarity, we design self-supervised loss functions tailored to radar characteristics. Extensive experiments on the K-Radar dataset demonstrate both qualitative and quantitative superiority of RadarMP for radar-based motion perception. We believe our work could inspire and advance future research on 4D imaging radar in motion perception for autonomous driving.
\section{Acknowledgment}
This project was supported by the National Natural Science Foundation of China (NSFC) under Grant No.62172043.

\bibliography{aaai2026}

\appendix
\renewcommand\thefigure{A\arabic{figure}}
\setcounter{figure}{0}
\renewcommand\thetable{A\arabic{table}}
\setcounter{table}{0}
\renewcommand\theequation{A\arabic{equation}}
\setcounter{equation}{0}
\pagenumbering{arabic}
\renewcommand*{\thepage}{A\arabic{page}}

\twocolumn[
 \vskip 0.520in minus 0.125in%
\centering%
{\LARGE\bf RadarMP: Motion Perception for 4D mmWave Radar in Autonomous Driving \\ \textit{(Appendix)} \par}%
\vskip 0.519in plus 0.5fil minus 0.05in%
]
\begin{figure*}[htbp]
	\centering
	{\includegraphics[width=\linewidth, trim=0 0 0 0, clip, keepaspectratio]{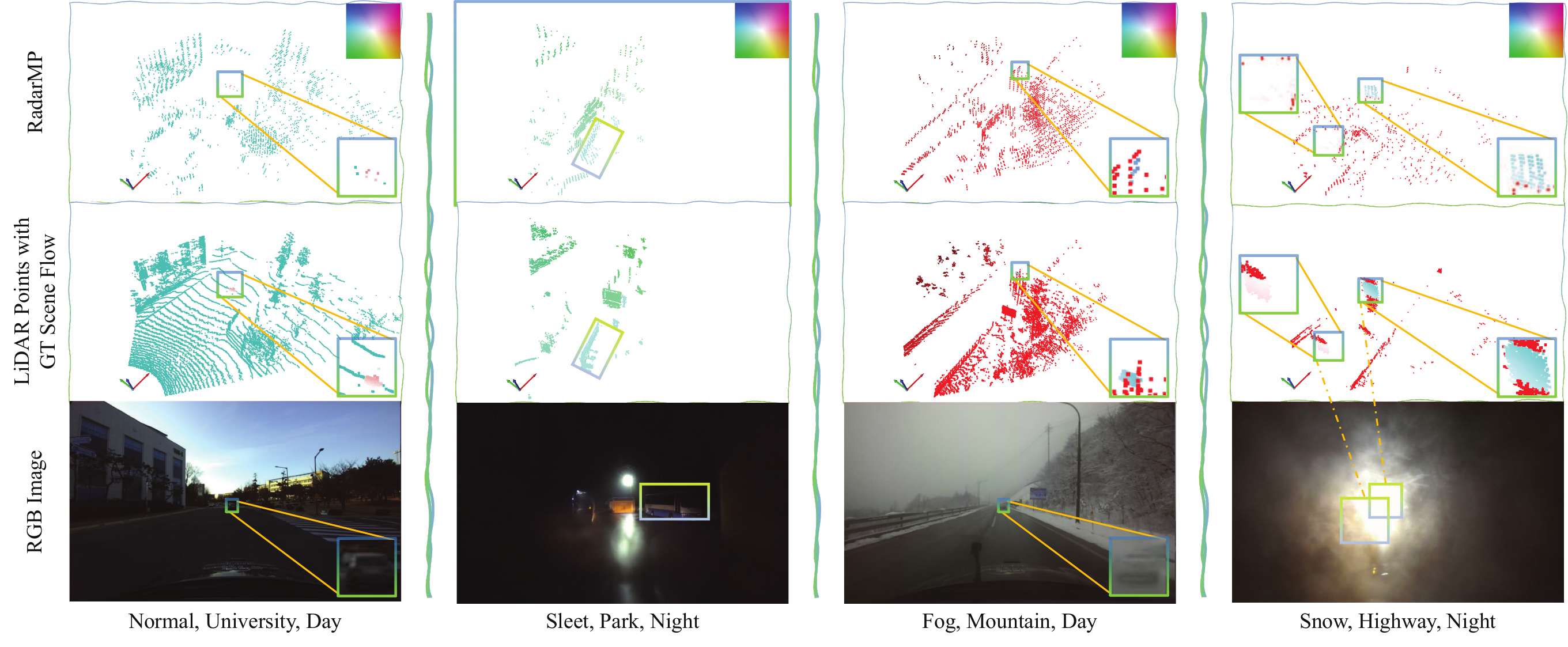}}
	\caption{\textbf{Qualitative results under different scenes.}  RadarMP achieves excellent performance across four distinct scenes. In particular, under heavy snow where both camera and LiDAR modalities experience severe degradation, our method continues to deliver reliable target detection and motion estimation, highlighting its potential to improve the safety of autonomous driving in adverse environments. The dynamic objects in the scene have been zoomed in at the bottom right to highlight the accuracy of non-rigid motion estimation. Colors indicate motion vectors in the XY plane only, scene conditions are shown at the bottom, and the RGB image is used for visualization only.}
	\label{result_supp}
\end{figure*}

\section{Related Work}
\subsection{Radar Target Detection Enhancement}
Cheng et al. \cite{rpdnet} proposed a radar detector network (RPDNet) that is supervised on the 2D signals in the Range-Doppler domain using pseudo-labels generated from LiDAR for enhancing mmWave radar point clouds. Recent studies by Prabhakara et al., Wu et al., and Zhang et al. \cite{radarhd, diffradar, diffral} treat radar point enhancement as a supervised image restoration task, leveraging LiDAR-based labels to denoise 2D radar heatmaps and extract valid BEV occupancy. 
Prabhakara et al. \cite{radarhd} approached radar point cloud enhancement as an image super-resolution task and proposed RadarHD. This network uses a U-net backbone, employing LiDAR to improve the resolution of radar heatmaps in the range-azimuth domain, and eliminates ghosting and reflections in the radar heatmaps. 
Wu et al. \cite{diffradar} proposed enhancing radar point using a diffusion probability model. To improve the decoding performance of the diffusion probability model, they utilized spatial attention and contour encoders to suppress multipath noise and extract contour information from the heatmap.
Inspired by image deblurring techniques, Zhang et al. \cite{diffral} also used diffusion models to refine radar 2D images and employed consistency models \cite{consmodels} to accelerate model inference, enabling their model's application in the autonomous navigation of micro aerial vehicles (MAVs).
Roldan et al. \cite{radelft} employed a radar detector network supervised by pre-processed 3D LiDAR voxel data and a Temporal Coherence Network to maintain temporal correlation across the radar frames.
\subsection{Scene Flow Estimation}
Junhwa et al. \cite{selfmonosf} were the first to employ a single network for monocular scene flow estimation, reducing model complexity while achieving competitive accuracy compared to multi-task approaches.
Teed et al. \cite{raft3d} extended RAFT \cite{raft}, which utilizes an iterative approach to estimate per-pixel 3D motion using dense SE3.
Liu et al. \cite{flownet3d} proposed the first end-to-end model for scene flow estimation based on point clouds.
PV-RAFT \cite{pvraft} combined the advantages of per-point and voxel-based correlations to refine local point correlations and global voxel correlations. 
PointPWC-Net \cite{pointpwcnet} firstly introduced the self-supervised loss for point cloud scene flow estimation. 
SLIM \cite{slim} exploited motion segmentation cues to develop a self-supervised loss that distinguishes between rigid and non-rigid scene flows.

There are few studies on scene flow estimation for mmWave radar point clouds. Ding et al. \cite{raflow, cmflow} leverage the unique Doppler information of mmWave radar and the redundant multimodal perception in autonomous vehicles for self-supervised and cross-modal supervision.
\section{Experimental Setup Details}
\subsection{Dataset Statitics}
We select 15 sequences from the K-Radar dataset \cite{kradar} covering diverse weather, lighting conditions, and scene types. These sequences are split into training, validation, and test sets with a ratio of 8:2:5, resulting in 4,768, 1,192, and 1,788 frames, respectively.
Under adverse weather conditions such as snow, rain, and fog, water droplets and snowflakes scatter or absorb laser beams, significantly degrading LiDAR performance and introducing noisy measurements. Therefore, for segmentation evaluation, which relies on LiDAR points to determine correctness, we exclude test sequences recorded during freezing rain and light snow conditions.
From the validation and test sets, we collect over 0.5 billion voxel-level samples based on annotated target detection and scene flow labels, with approximately 96.80\% labeled as noise, 2.91\% as rigid targets, and 0.29\% as non-rigid targets.
\subsection{Ground Truth Labelling}
We annotate labels only for the validation and test sets.
\subsubsection{Target Detection Annotation}
As discussed in the main text, radar target detection is still approached as a segmentation task. Segmentation labels are derived from LiDAR point clouds. First, LiDAR points are filtered to match the radar field of view (FOV) and projected onto a Cartesian grid aligned with the radar spatial resolution, resulting in a voxel grid with the same spatial dimensions as the input radar tesseract.
To balance the resolution difference between LiDAR and radar, a voxel is labeled as a radar target if the neighborhood around its center with a radius of 0.5m contains more than three LiDAR points, while others are discarded as noise.
\subsubsection{Scene Flow Annotation}
K-Radar provides odometry information and 3D object detection and tracking annotations for each sequence. We follow \cite{sceneflowanno, cmflow} to generate scene flow labels. Specifically, we decompose the scene motion field into rigid and non-rigid motions. Rigid motion $T_{s}$ refers to the ego-motion of the radar, i.e., the apparent motion of static objects in the radar coordinate frame. $T_{s}$ is obtained using the relative transformation between the two-frame odometry poses and the radar extrinsics. Non-rigid motion refers to the actual movement of dynamic objects in the scene. Using the provided 3D detection and tracking annotations, we estimate the relative pose $T_{d}$ of each annotated bounding box of a unique object across two frames and determine whether it is static or dynamic. The scene flow of the target point is computed as the difference between its point-wise coordinates before and after transformation using the corresponding translation matrix ($T_{s}$ or $T_{d}$).
\subsection{License}
The K-Radar dataset is published under the CC BY-NC-ND License, and all K-Radar codes are published under the Apache License 2.0.

\section{Implementation Details of RadarMP}
\subsection{Data Preprocessing}
To balance detection performance and memory efficiency, we apply dimension reduction to the original radar tesseract. The original tesseract volume has a shape of $64\times 256\times 107\times 37$, corresponding to Doppler, range, azimuth, and elevation dimensions, respectively.
We truncate the range dimension to the first 128 bins to focus on the core sensing region. The azimuth dimension is symmetrically clipped to 96 bins and further downsampled to 48 bins by averaging. The elevation dimension is symmetrically clipped to 32 bins.
As a result, the final input size is $64\times 128\times 48\times 32$ ($D\times R\times A\times E$), and the polar coordinate ranges for the three spatial dimensions approximately are [0m,0.46m,128m], [-47\textdegree,1\textdegree,48\textdegree], [-15\textdegree,1\textdegree,16\textdegree], representing the minimum value, resolution, and maximum value of each dimension.

\subsection{Correlation Reference Points}
As discussed in the main text, the reference coordinates $\mathbf{p}_l\in \mathbb{R}^{3 \times \frac{R}{2^l} \times \frac{A}{2^l} \times \frac{E}{2^l}}$ are obtained by averaging the 2D flow prediction from each dimension, as defined by the following equation:
\begin{equation}
\begin{aligned}
r'_l &= r_l + \frac{1}{2}\left(\mathbf{f}^l_{ra}(r) + \mathbf{f}^l_{re}(r)\right)\text{,} \\
a'_l &= a_l + \frac{1}{2}\left(\mathbf{f}^l_{ra}(a) + \mathbf{f}^l_{ae}(a)\right)\text{,} \\
e'_l &= e_l + \frac{1}{2}\left(\mathbf{f}^l_{re}(e) + \mathbf{f}^l_{ae}(e)\right)\text{,} \\
\mathbf{p}_l  &= \mathrm{concat}(r'_l,~a'_l,~e'_l)\text{,} \\
\end{aligned}
\label{warprefpoint}
\end{equation}
where $\mathbf{p}_l\in \mathbb{R}^{3 \times \frac{R}{2^l} \times \frac{A}{2^l} \times \frac{E}{2^l}}$ denotes the reference point coordinates at $l_{th}$ level,  $r_l$ denotes the origin range coordinate index of the $l_{th}$-level query feature, $r'_l$ represents the range coordinate index of its corresponding reference point in the value feature.
We extend each 2D flow field along its missing spatial dimension to form 3D flow volume, enabling $\mathbf{f}^l_{ra}(r)\in \mathbb{R}^{\frac{R}{2^l} \times \frac{A}{2^l} \times \frac{E}{2^l}}$ to denote the range dimension slice of the $l_{th}$-level flow volume in $\mathbf{F}_{ra}$.
In addition, the pretrained PWC-Net is further fine-tuned using 2D planar radar data, with the loss derived from a simplified 2D version of the Energy Flow Loss (i.e., Eq.~\eqref{efl} in the main text).
\subsection{Training Details}
We train RadarMP with 250 epochs, using a batch size of 3 per GPU.
As discussed in the main text, we optimize the network using three loss functions specifically designed based on radar characteristics. To balance their weights, we follow \cite{MDGA} to perform loss normalization. Additionally, we adopt a cosine annealing strategy with a warm-up ratio of 1/5 during the initial training phase.
\subsection{Effciency Report}
RadarMP achieves 7.6 fps inference on a 3090 using
7.5 GB GPU memory. When reducing the Doppler dimension to 32/16 (top-K bins), storage usage is alleviated, while $P_d$ drops 3.6\%/4.3\% and EPE3D drops 0.011$m$/0.018$m$.
\section{Additional Experiment Results}
We present additional qualitative results on the remaining test sequences in Figure~\ref{result_supp}, covering diverse scenes and four distinct weather conditions, with notable variations in illumination. As shown, RadarMP delivers reliable performance across all weather and lighting scenarios, which is attributed to our unique design that enforces motion consistency during detection, allowing radar as a standalone modality to produce stable and reliable motion perception. This capability is particularly valuable when optical sensors degrade due to low illumination conditions or adverse weather, enabling adequate perceptual compensation.

However, limitations remain. Due to limited radar resolution and the presence of noise, our method struggles in scenarios involving low Radar Cross Section (RCS) targets adjacent to clutter. In such cases, the direction of energy flow cannot be reliably inferred across frames, leading to segmentation ambiguity and inaccurate scene flow predictions.

\section{Limitations and Future Works}
Due to the inherent resolution constraints of radar, our work cannot provide the rich texture information available from LiDAR. The 4D radar tesseract also contains substantial noise, which poses challenges to accurate inter-frame motion estimation and limits its precision compared to LiDAR-based methods. Furthermore, the lack of precise radar point-level annotations hinders a more detailed analysis of target detection performance for RadarMP. Future work will aim to address these issues to improve motion perception accuracy further.

\end{document}